\documentclass[10pt,twocolumn,letterpaper]{article}

\usepackage{cvpr}
\usepackage{times}
\usepackage{epsfig}
\usepackage{graphicx}
\usepackage{amsmath}
\usepackage{amssymb}
\usepackage{CJK}
\usepackage{color}
\usepackage{booktabs,threeparttable}
\usepackage{colortbl}
\usepackage[export]{adjustbox}
\usepackage{makecell}
\usepackage{caption}

\usepackage[pagebackref=true,breaklinks=true,letterpaper=true,colorlinks,bookmarks=false]{hyperref}

\cvprfinalcopy 

\def\cvprPaperID{} 

\begin{document}

\title{Designing Network Design Strategies Through Gradient Path Analysis}

\author{Chien-Yao Wang$^{1}$, Hong-Yuan Mark Liao$^{1}$, and I-Hau Yeh$^{2}$ \\
	$^{1}$Institute of Information Science, Academia Sinica, Taiwan\\
	$^{2}$Elan Microelectronics Corporation, Taiwan\\
	{\tt\small kinyiu@iis.sinica.edu.tw, liao@iis.sinica.edu.tw, and ihyeh@emc.com.tw}
}

\maketitle

\begin{abstract}
	\vspace{-2mm}
	Designing a high-efficiency and high-quality expressive network architecture has always been the most important research topic in the field of deep learning.  Most of today's network design strategies focus on how to integrate features extracted from different layers, and how to design computing units to effectively extract these features, thereby enhancing the expressiveness of the network.  This paper proposes a new network design strategy, i.e., to design the network architecture based on gradient path analysis.  On the whole, most of today's mainstream network design strategies are based on feed forward path, that is, the network architecture is designed based on the data path.  In this paper, we hope to enhance the expressive ability of the trained model by improving the network learning ability.  Due to the mechanism driving the network parameter learning is the backward propagation algorithm, we design network design strategies based on back propagation path. We propose the gradient path design strategies for the layer-level, the stage-level, and the network-level, and the design strategies are proved to be superior and feasible from theoretical analysis and experiments.
\end{abstract}

\vspace{-6mm}
\section{Introduction}
\label{sec:intro}

Deep Neural Networks (DNNs) are now widely used on a variety of devices to solve different kinds of tasks.  Millions of scientists, engineers, and researchers are involved in deep learning-related work. They all look forward to designing efficient, accurate, low-cost solutions that can meet their needs.  Therefore, how to design network architectures suitable for their products becomes particularly important.

Since 2014, many DNNs have achieved near-human or superior performance than humans on various tasks.  For example, Google's GoogLeNet \cite{szegedy2015going} and Microsoft's PReLUNet \cite{he2015delving} on image classification, Facebook's Deepface \cite{taigman2014deepface} on face verification, and DeepMind's AlphaGo \cite{silver2016mastering} on the Go board, etc.  Based on the beginning of the above fields, some researchers continue to develop new architectures or algorithms that are more advanced and can beat the above methods; other researchers focus on how to make DNN-related technologies practical in the daily life of human beings.  SqueezeNet \cite{iandola2016squeezenet} proposed by Iandola \etal is a representative example, because it reduces the number of parameters of AlexNet \cite{krizhevsky2017imagenet} by 50 times, but can maintain a comparable accuracy.  MobileNet \cite{howard2017mobilenets, sandler2018mobilenetv2, howard2019searching} and ShuffleNet \cite{zhang2018shufflenet, ma2018shufflenetv2} are also good examples.  The former adds the actual hardware operating latency directly into the consideration of the architecture design, while the latter uses the analysis of hardware characteristics as a reference for designing the neural network architecture.

\begin{figure*}[t]
	\begin{center}
		\includegraphics[width=0.9\linewidth]{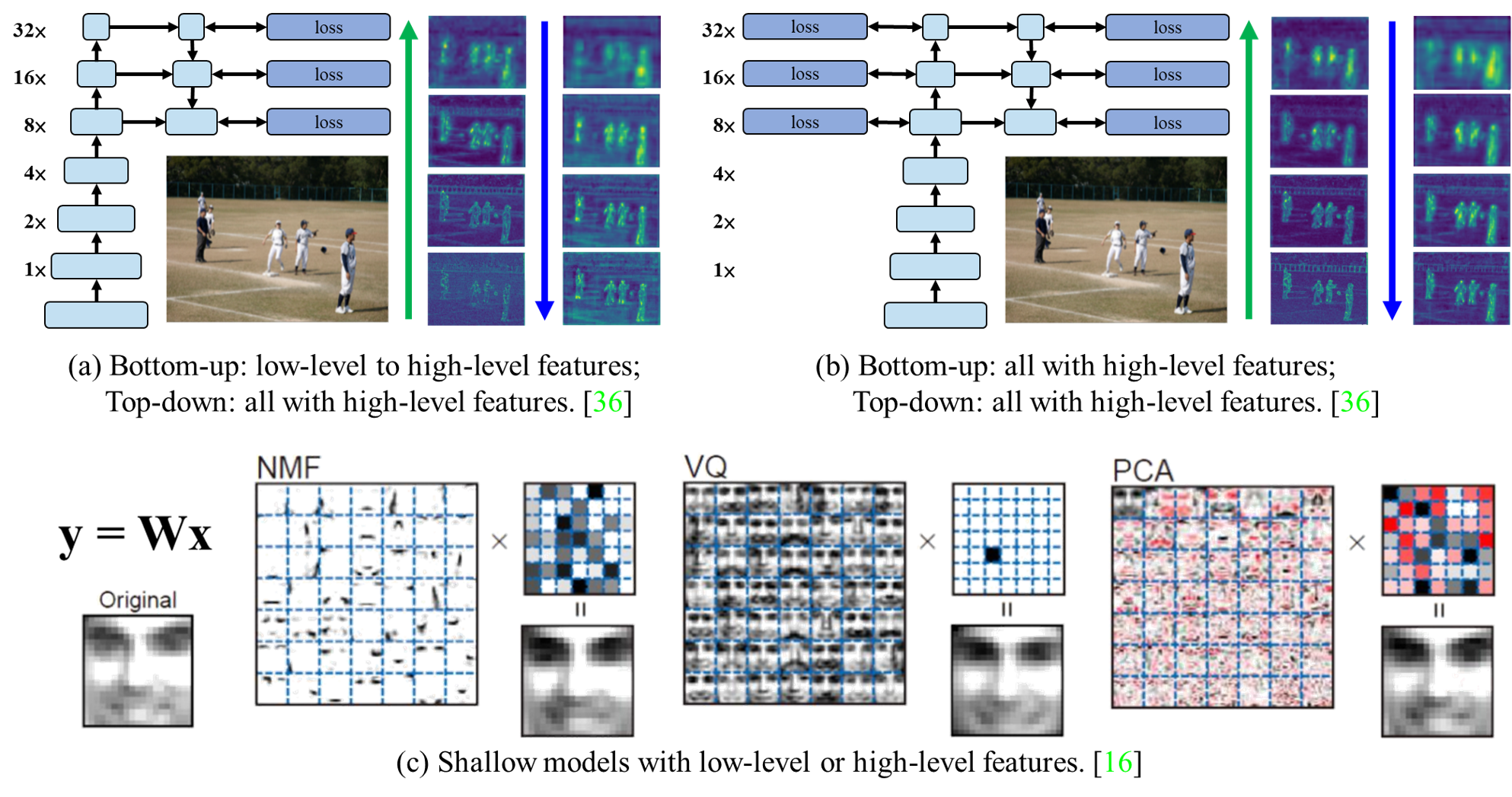}
	\end{center}
	\caption{We can find that no matter for shallow or deep models, and for shallower layers or deeper layers in a deep network, all of them have abilities to extract low-level or high-level features. 
	}
	\label{fig:observation}
	\vspace{-4mm}
\end{figure*}

Just after the ResNet \cite{he2016deep}, ResNeXt \cite{xie2017aggregated}, and DenseNet \cite{huang2017densely} architectures solved the convergence problem encountered in ultra-deep network training, the design of CNN architecture in recent years has focused on the following points: (1) feature fusion, (2) receptive field enhancement, (3) attention mechanism, and (4) branch selection mechanism.  In other words, most studies follow the common perception of deep networks, i.e., extract low-level features from shallow layers and high-level features from deep layers.  According to the above principles, one can use them to design neural network architectures to effectively combine different levels of features in data path (feed forward path.)  However, is such a design strategy necessarily correct?  We therefore analyze \cite{lee1999learning, yang2019dually}, articles that explore the difference in feature expression between shallow and deep model using different objectives and loss layers.  From Figure \ref{fig:observation}, we found that by adjusting the configuration of objectives and loss layers, we can control the features learned by each layer (shallow or deep).  That is to say, what kind of features the weight learns is mainly based on what kind of information we use to teach it, rather than the combination of those layers that the input comes from.  Based on this finding, we redefine network design strategies.

Since we propose that the network architecture can be designed with the concept that objective function can guide neural network to learn information, we must first understand how an objective function affects the update of network weights.  At present, the main weight update method is the backpropagation algorithm, which uses partial differentiation to generate gradients, and then updates the weights by gradient decent.  This algorithm propagates gradient information to the shallow layers in chain rule manner, and repeats such steps until the weights of all layers are updated.  In other words, the information that an objective function teaches is propagated between layers in the form of gradients.  In this paper, we propose that by analyzing the gradient generated through the guidance of objective function, we can design the network architecture by the gradient paths when executing the backpropagation process.  We design the network architecture for three different levels of strategies such as layer-level design, stage-level design, and network-level design, which are described below:
\begin{enumerate}
	\item \textbf{Layer-level design:} At this level we design gradient flow shunting strategies and use them to confirm the validity of the hypothesis.  We adjust the number of layers and calculate the channel ratio of residual connection, and then design Partial Residual Network (PRN) \cite{wang2019enriching}, as described in Section \ref{sec:prn}.
	
	\vspace{-1mm}
	
	\item \textbf{Stage-level design:} We add hardware characters to speed up inference on the network.  We maximize gradient combinations while minimizing hardware computational cost, and thus design Cross Stage Partial Network (CSPNet) \cite{wang2020cspnet}, as described in Section  \ref{sec:cspn}.	
	
	\vspace{-1mm}
	
	\item \textbf{Network-level design:} We add the consideration of gradient propagation efficiency to balance the leaning ability of the network.  When we design the network architecture, we also consider the gradient propagation path length of the network as a whole, and therefore design Efficient Layer Aggregation Network (ELAN), as described in Section \ref{sec:elan}.
\end{enumerate}


\begin{figure*}[t]
	\begin{center}
		\includegraphics[width=0.75\linewidth]{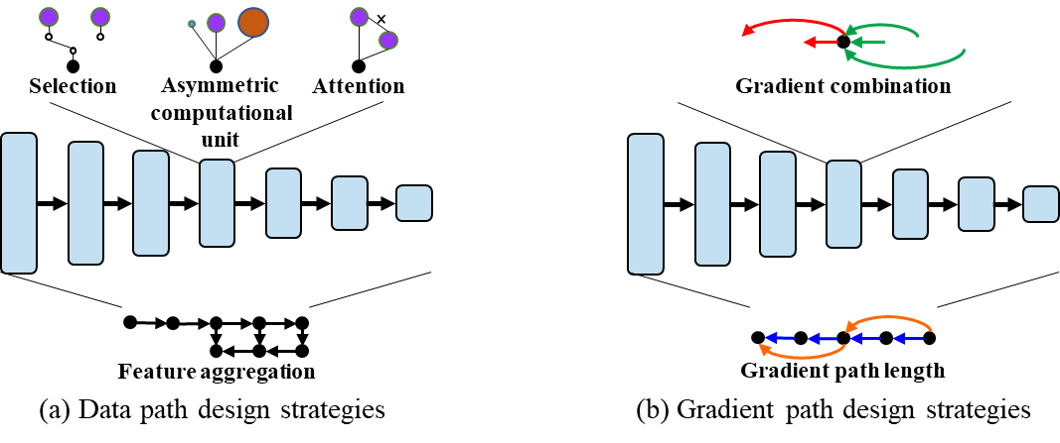}
	\end{center}
	\caption{Two main network design strategies: (a) data path design strategy; and (b) gradient path design strategy.}
	\label{fig:strategy}
	\vspace{-4mm}
\end{figure*}

\section{Methodology}
\label{sec:method}

\subsection{Network Design Strategies}
\label{sec:nds}

In this paper we divide network design strategies into two kinds: (1) data path design strategies, and (2) gradient path design strategies, as shown in Figure \ref{fig:strategy}.  Data path design strategy mainly focuses on designing feature extraction, feature selection, and feature fusion operations to extract features with specific properties.  These features can help subsequent layers use these features to further obtain better properties for conducting more advanced analysis.  The purpose of applying gradient path design strategies is to analyze the source and composition of the gradients, and how they are updated by the driving parameters.  Then, one can use the results of the above analysis to design the network architecture.  The design concept is to hope that the final parameter utilization rate is higher, and thereby achieve the best learning effect.

Next, we will discuss the advantages and disadvantages of the data path design strategy and the gradient path design strategy, respectively.  There are three advantages of the data path design strategy: (1) \textbf{can extract features with specific physical meaning.}  For example, use asymmetric computational units to extract features with different receptive fields \cite{he2015spatial, chen2017deeplab, liu2018receptive}; (2) \textbf{can automatically select suitable operation units with parameterized models for different inputs.}  For example, using kernel selection to handle inputs with different properties \cite{li2019selective, chen2020dynamic}; and (3) \textbf{the learned features can be reused directly.}  For example, feature pyramid networks can directly utilize features extracted from different layers for more accurate predictions \cite{lin2017feature}.  The data path design strategy has two shortcomings: (1) In the process of training, it sometimes leads to unpredictable degradation of the effect, and at this time, a more complex architecture needs to be designed to solve the problem.  For example, the pairwise relationship of non-local networks is easy to degenerate into unary information \cite{yin2020disentangled}; and (2) various specially designed arithmetic units are easy to cause difficulties in performance optimization.  For example, in the ASIC design dedicated to AI, if the designer wants to add an arithmetic unit, an additional set of circuits is required.

As for the gradient path design strategy, there are in total three advantages: (1) \textbf{can effectively use network parameters.}  In this part, we propose that by adjusting the gradient propagation path, the weights of different computing units can learn various information, and thereby achieve higher parameter utilization efficiency; (2) \textbf{has stable model learning ability.}  Since gradient path design strategy directly determines and propagates information to update weights to each computing unit, the designed architecture can avoid degradation during training; and (3) \textbf{has efficient inference speed.}  The gradient path design strategy makes parameter utilization very efficient, so the network can achieve higher accuracy without adding additional complex architecture.  Because of the above reasons, the designed network can be lighter and simpler in architecture.  The proposed gradient path design strategy only has one shortcoming, i.e., when the gradient update path is not a simple reversed feedforward path of the network, the difficulty of programming will be greatly increased.

\begin{figure}[h]
	\vspace{4mm}
	\begin{center}
		\includegraphics[width=1.0\linewidth]{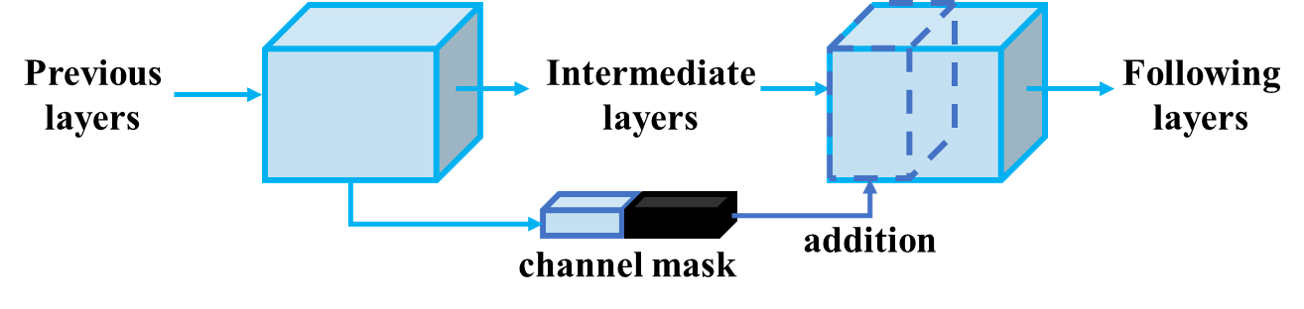}
	\end{center}
	\caption{Masked residual layer. Only part of channels will go through the identity connection.}
	\label{fig:mres}
\end{figure}

\subsection{Partial Residual Networks}
\label{sec:prn}

Partial Residual Network (PRN) \cite{wang2019enriching} was proposed by our team in 2019, and its design concept belongs to the layer-level design strategy.  In the design of PRN, the main concept is to maximize the combination of gradients used to update the weights of each layer.  There are two main factors that affect the combination of gradients.  The first one is \textbf{the source layer of the gradient}.  The source layer is composed of the nodes connected the indegree edges of the gradient path.  The second factor that affects gradient combination is \textbf{the time it takes for the gradient flow to arrive at a particular layer from the loss layer through the operation of the chain rule}.  One thing to be noted is that when the gradient changes during the process of the chain rule update, the amount of loss information it covers will gradually fade as the chain grows.  We define the above time duration as the number of layers that the gradient flow needs to travel from the loss layer to a specific layer.  In PRN, we propose the following two structures to improve ResNet:

\textbf{Masked residual layer.} In the design of ResNet \cite{he2016deep}, the output of each computational block is added together with an identity connection, and such a structure is called residual layer.  In PRN, we multiply identity connection by a binary mask and only allow the feature map of some of the channels to be added to the output of the computational block.  We call this structure masked residual layer, and its architecture is shown in Figure  \ref{fig:mres}.  Using the mechanism of a masked residual layer allows the feature map to be divided into two parts, in which the weights corresponding to the channels that are masked and the weights corresponding to the channels with identity connection will significantly increase the number of gradient combinations due to the aforementioned masking effect.  In addition, differences in gradient sources will simultaneously affect the overall gradient timestamp (time node along time axis), thus making gradient combinations more abundant. 

\newpage

\textbf{Asymmetric residual layer.} Under the ResNet architecture, only feature map of the same size can be added, which is why it is a very restricted architecture.  Generally, when the calculation amount and inference speed of the optimized architecture are performed, we are often limited by this architecture and cannot design an architecture that meets the requirements.  Under the architecture of PRN, the masked residual layer proposed by us can regard the inconsistency of the number of channels as some channels being blocked, and thus allow feature map with different number of channels to perform masked residual operations.  We call the layer that operates in the above manner an asymmetric residual layer.  An asymmetric residual layer is designed in such a way that the network architecture is more flexible and more able to maintain the properties of a gradient path-based model.  For example, when we are doing feature integration, the general approach requires additional transition layers to project different feature maps to the same dimension, and then perform the addition operation.  However, the above-mentioned operation will increase a large number of parameters and amount of computations, and will also make the gradient path longer, and thus affect the convergence of the network.  The introduction of asymmetric residual layer can perfectly solve similar issues.

\subsection{Cross Stage Partial Networks}
\label{sec:cspn}

CSPNet \cite{wang2020cspnet} was proposed by our team in 2019, and it is a stage-level gradient path-based network. Like PRN, CSPNet is based on the concept of maximizing gradient combinations. The difference between CSPNet and PRN is that the latter focuses on confirming the improvement of network learning ability by gradient combination from theoretical perspective, while the former is additionally designed for further architecture optimization for hardware inference speed. Therefore, when designing CSPNet, we extend the architecture from layer-level to stage-level, and optimize the overall architecture. CSPNet mainly has the following two structures:

\begin{figure}[h]
	\begin{center}
		\includegraphics[width=1.0\linewidth]{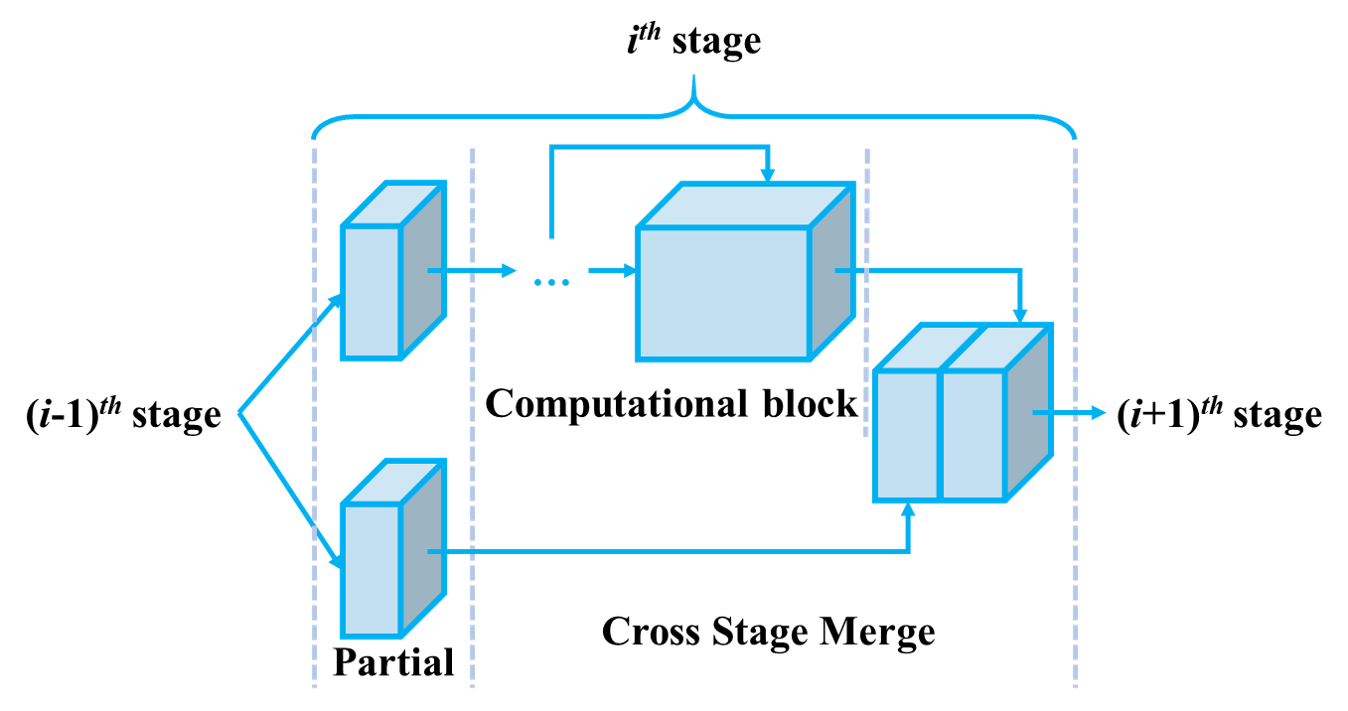}
	\end{center}
	\caption{Cross stage partial operation. CSP operation
		separates feature map of the base layer into two parts, one part will go through a computational block and the other part is then combined with processed feature map to the next stage.}
	\label{fig:csp}
\end{figure}

\newpage

\textbf{Cross stage partial operation.} From the perspective of maximizing the source of the gradient, we can easily find that the source of the gradient can be maximized when each channel has a different gradient path. Also, from the perspective of maximizing gradient timestamps, we know that the number of gradient timestamps can be maximized when each channel has computational blocks of different depths. Following the above concept, we can derive an architecture designed to maximize both the gradient source and gradient timestamp. And this architecture will be the Inception-like architecture \cite{szegedy2015going, ioffe2015batch, szegedy2016rethinking, szegedy2017inception} and the fractal-like architecture \cite{larsson2017fractalnet} with depth-wise convolution. Although the above design can effectively improve the parameter utilization, it will greatly reduce the parallelization ability. In addition, it will cause the model to significantly reduce the inference speed on inference engines such as GPU and TPU. From the previous analysis, we know that dividing the channel can increase the number of gradient sources, and making the sub-networks connected by different channels with different layers can increase the number of gradient timestamps. The cross stage partial operation we designed can maximize the combination of gradients and increase the inference speed without breaking the architecture and can be parallelized.  This architecture is shown in Figure \ref{fig:csp}. In Figure \ref{fig:csp}, we divide a stage's input feature map into two parts, and use this manner to increase the number of gradient sources. The detailed procedure is as follows: we first divide the input feature map into two parts and one of them passes through the computational block, and this computational block can be any computational block such as Res block, ResX block, or Dense block. As for the other part, it directly crosses the entire stage, and then integrates with the part that goes through the computational block. Since only part of the feature map enters the computational black for operation, this kind of design can effectively reduce the amount of parameters, operation, memory traffic, and memory peak, allowing the system to achieve faster inference speed.

\begin{figure*}[t]
	\begin{center}
		\includegraphics[width=.95\linewidth]{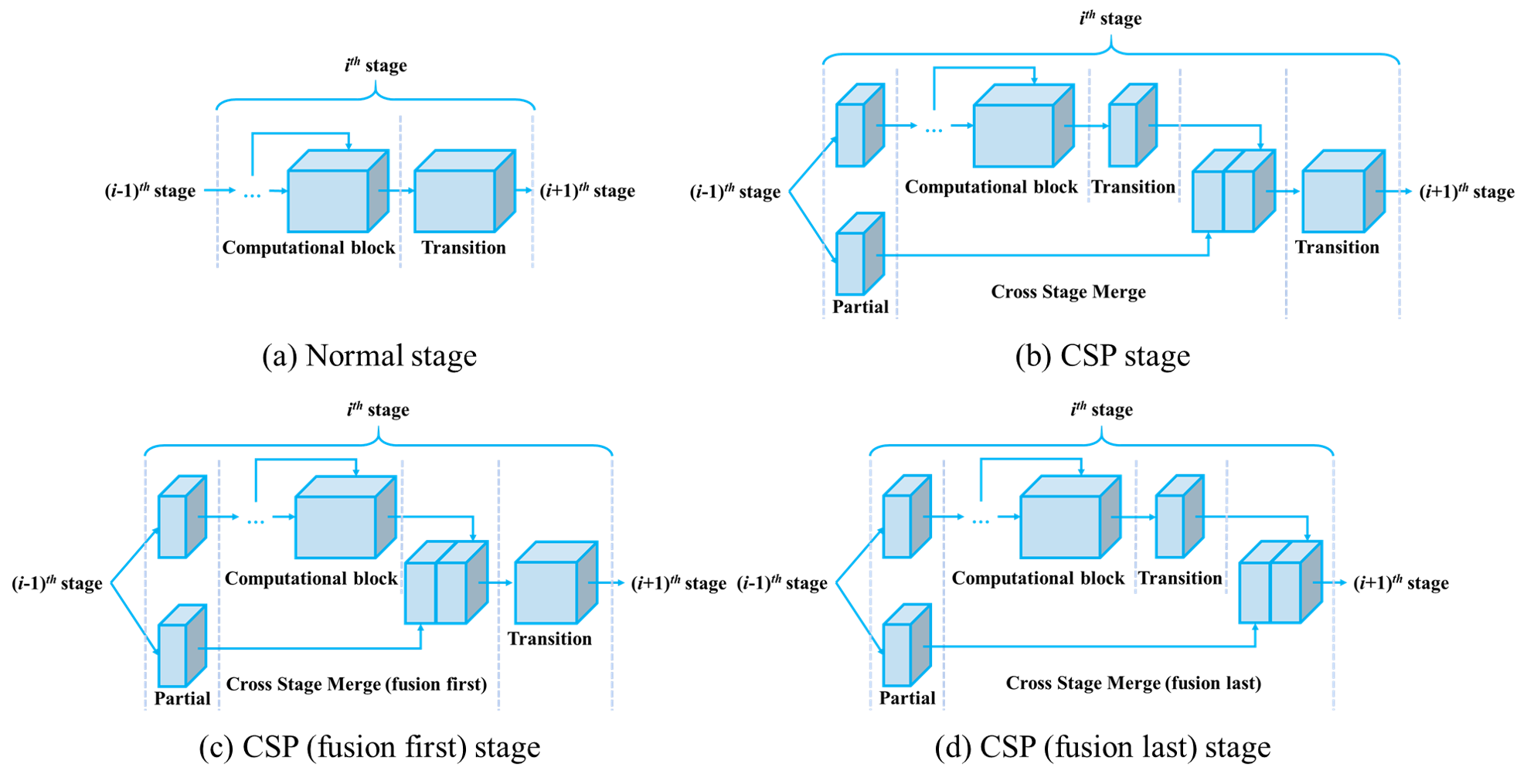}
	\end{center}
	\caption{Cross stage partial networks. (a) original network, (b) CSP fusion: transition $\rightarrow$ concatenation $\rightarrow$ transition, (c) CSP fusion first: concatenation $\rightarrow$ transition, and (d) CSP fusion last: transition $\rightarrow$ concatenation}
	\label{fig:cspn}
	\vspace{-4mm}
\end{figure*}

\textbf{Gradient flow truncate operation.} In order to make our designed network architecture more powerful, we further analyze the gradient flow used to update the CSPNet.  Since shortcut connections are often used in computational blocks, we know that the gradient sources that provide the two paths are bound to overlap a lot. We know that when a feature map passes through a kernel function, it is equivalent to a spatial projection.  Usually we can insert a transition layer at the end of both paths to truncate the duplicated gradient flow.  Through the above steps, we can make the information learned from the two paths and adjacent stages have more obvious diversity.  We designed three different combinations of duplicate gradient flow truncate operations, as shown in Figure \ref{fig:cspn}.  These operations can be matched with different architectures, such as computational blocks and down-sampling blocks to achieve better results.

\subsection{Efficient Layer Aggregation Networks}
\label{sec:elan}

The codes of \textbf{Efficient Layer Aggregation Networks (ELAN)} was released by our team in July 2022.  It falls into the category of the \textbf{gradient path designed network} at the network-level.  The main purpose of designing ELAN is to solve the problem that the convergence of the deep model will gradually deteriorate when executing model scaling.  We analyze the shortest gradient path and the longest gradient path through each layer in the overall network, thereby designing a layer aggregation architecture with efficient gradient propagation paths.  ELAN is mainly composed of VoVNet \cite{lee2019energy} combined with CSPNet \cite{wang2020cspnet}, and optimizes the gradient length of the overall network with the structure of \textbf{stack in computational block}.  In what follows, we will elaborate how stack in computational block works.

\begin{figure*}[t]
	\begin{center}
		\includegraphics[width=.85\linewidth]{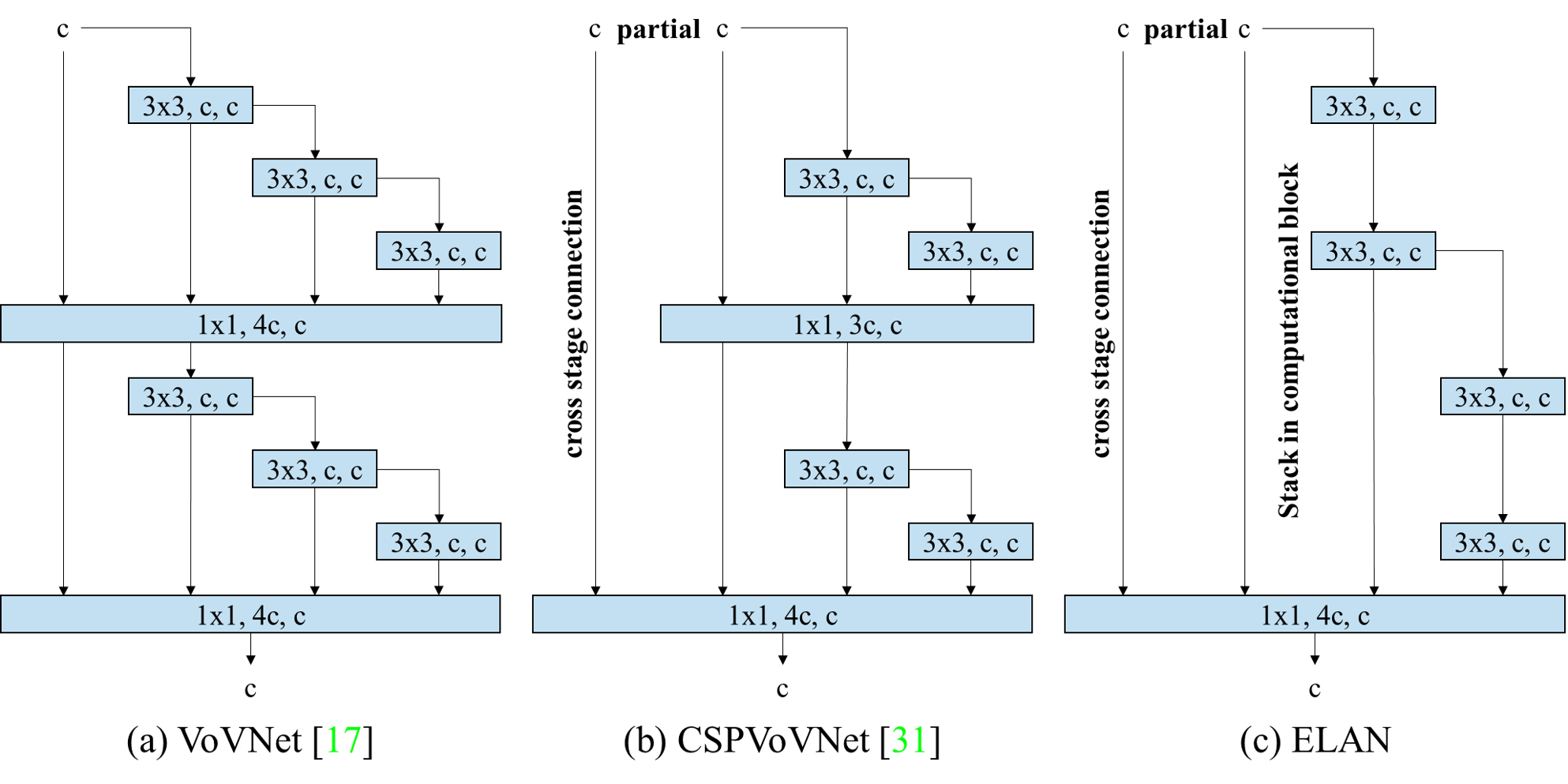}
	\end{center}
	\caption{Efficient layer aggregation networks.}
	\label{fig:elan}
	\vspace{-4mm}
\end{figure*}

\textbf{Stack in computational block.} When we are doing model scaling, there will be a phenomenon, that is, when the network reaches a certain depth, if we continue to stack computational blocks, the accuracy gain will be less and less.  To make matters worse, when the network reaches a certain critical depth, its convergence begins to deteriorate, resulting in an overall accuracy that is worse than shallow networks.  One of the best examples is scaled-YOLOv4 \cite{wang2021scaled}, we see that its P7 model uses expensive parameters and operations, but only a small amount of accuracy gain, and the same phenomenon occurs in many popular networks.  For example, ResNet-152 is about three times as computationally intensive as ResNet-50, but offers less than 1\% improvement in accuracy on ImageNet \cite{he2016deep}.  When ResNet is stacked to 200 layers, its accuracy is even worse than ResNet-152 \cite{he2016identity}.  Also, when VoVNet is stacked to 99 layers, its accuracy is even much lower than that of VoVNet-39 \cite{lee2019centermask}.  From the gradient path design strategy point of view, we speculate that the reason why the accuracy of VoVNet degenerates much faster than ResNet is because the stacking of VoVNet is based on the OSA module.  We know that every OSA module contains a transition layer, so every time we stack an OSA module, the shortest gradient path of all layers in the network increases by one.  As for ResNet, it is stacked by residual blocks, and the stacking of residual layers will only increase the longest gradient path, and will not increase the shortest gradient path.  In order to verify the possible effects of model scaling, we did some experiments based on YOLOR-CSP \cite{wang2021you}.  From the experimental results we found that when the stacking layer reaches 80+ layers, the accuracy of \textbf{CSP fusion first} starts to perform better than the normal CSPNet.  At this point, the shortest gradient path of the computational block of each stage will be reduced by 1.  As the network continues to widen and deepen, \textbf{CSP fusion last} will get the highest accuracy, but at this point the shortest gradient path of all layers will be reduced by 1.  The above experimental results confirmed our previous hypothesis.  With the support of the above experiments, we designed the ``stack in computational block'' strategy in ELAN, as shown in Figure \ref{fig:elan}.  The purpose of our design is to avoid the problem of using too many transition layers and making the shortest gradient path of the whole network quickly become longer.  We hope that the above design strategy allows ELAN to be successfully trained when the network is stacked deeper.

\section{Analysis}
\label{sec:analysis}

In this section we will analyze the proposed gradient path design strategies based on the classical network architecture.  First, we will analyze the existing network architecture and the proposed PRN with the concept of gradient combination, and this example shows that the network architecture that performs well does have a richer gradient combination. Then we will analyze how the proposed CSPNet brings richer gradient combinations and other benefits.  Finally, we analyze the importance of length of gradient path by stop gradient, and thus confirm that the proposed ELAN has a design concept advantage.

\begin{table}[t]
	\centering
	\begin{threeparttable}[t]
		\footnotesize
		\caption{Analysis of different networks. \cite{zhu2018sparsely}}
		\label{table:a1}
		\setlength\tabcolsep{4.5pt}
		\begin{tabular}{lccc}
			\toprule
			\textbf{Model} & \textbf{Parameters} & \thead{\textbf{Shortest} \\ \textbf{gradient path}} & \thead{\textbf{Aggrgated} \\ \textbf{features}}  \\				
			\midrule
			\textbf{PlainNet} & $O(N)$ & $O(N)$ & $O(1)$ \\
			\textbf{ResNet \cite{he2016deep}} & $O(N)$ & $O(1)$ & $O(l)$ \\		
			\textbf{DenseNet \cite{huang2017densely}} & $O(N^{2})$ & $O(1)$ & $O(l)$ \\
			\textbf{Sparse ResNet \cite{zhu2018sparsely}} & $O(N)$ & $O(\log N)$ & $O(\log l)$ \\
			\textbf{Sparse DenseNet \cite{zhu2018sparsely}} & $O(N \log N)$ & $O(\log N)$ & $O(\log l)$ \\
			\bottomrule
		\end{tabular}
	\end{threeparttable}
\end{table}

\subsection{Analysis of gradient combination}
\label{sec:agc}

General researchers often use the shortest gradient path and the number of integrated features to measure the learning efficiency and ability of network architectures.  However, from the literature \cite{zhu2018sparsely} we can find that these metrics are not completely related to accuracy and parameter usage, as shown in Table \ref{table:a1}.  We observe the process of gradient propagation and find that the gradient combination used to update the weights of different layers matches the learning ability of the network well, and in this section we will analyze the gradient combination.  Gradient combinations are composed by two types of component, namely \textbf{gradient timestamp} and \textbf{gradient source}.  Next we will analyze them separately.

\begin{figure*}[t]
	\begin{center}
		\includegraphics[width=.85\linewidth]{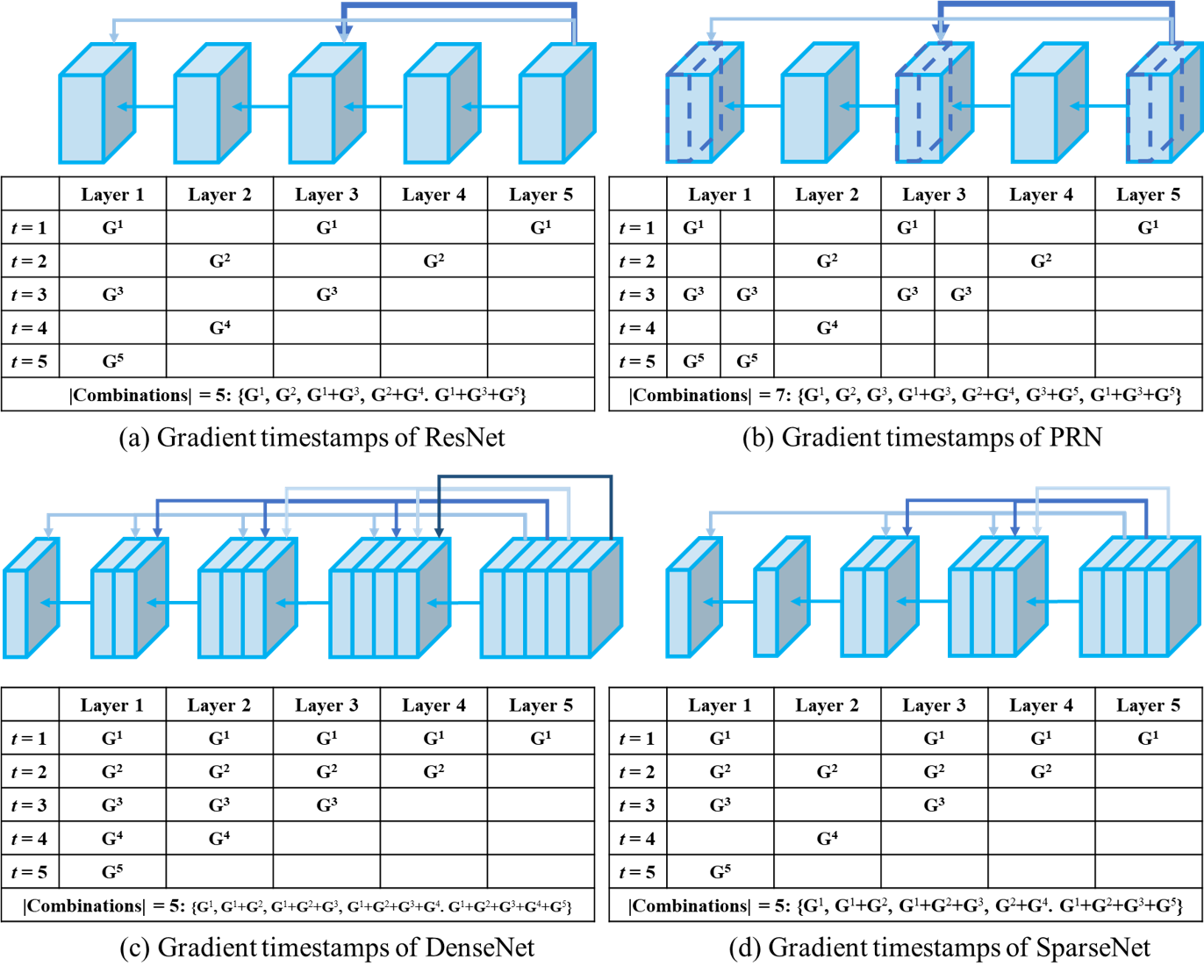}
	\end{center}
	\caption{Gradient timestamps. (a) ResNet, (b) PRN, (c) DenseNet, and (d) SparseNet.}
	\label{fig:time}
\end{figure*}

\textbf{Gradient Timestamp.} Figure \ref{fig:time} shows the architecture of ResNet \cite{he2016deep}, PRN, DenseNet \cite{huang2017densely}, and SparseNet \cite{zhu2018sparsely}.  Among them, we unfold the cascaded residual connection and concatenation connection to facilitate the observation of the gradient propagation process.  In addition, the gradient flow delivery timestamps on each architecture is also shown in Figure \ref{fig:time}.  The gradient sequence is equivalent to a breadth first search process, and each sequence will visit all the outdegree nodes reached by the previous round of traverse.  From Figure \ref{fig:time}, we can see that PRN uses the channel splitting strategy to enrich the gradient timestamps received by the weights corresponding to different channels.  As for SparseNet, it uses sparse connections to make the timestamps received by the weight connections corresponding to different layers more variable.  Both of the above methods can learn more diverse information with different weights, which makes our proposed architecture more powerful.

\begin{figure}[h]
	\begin{center}
		\includegraphics[width=1.0\linewidth]{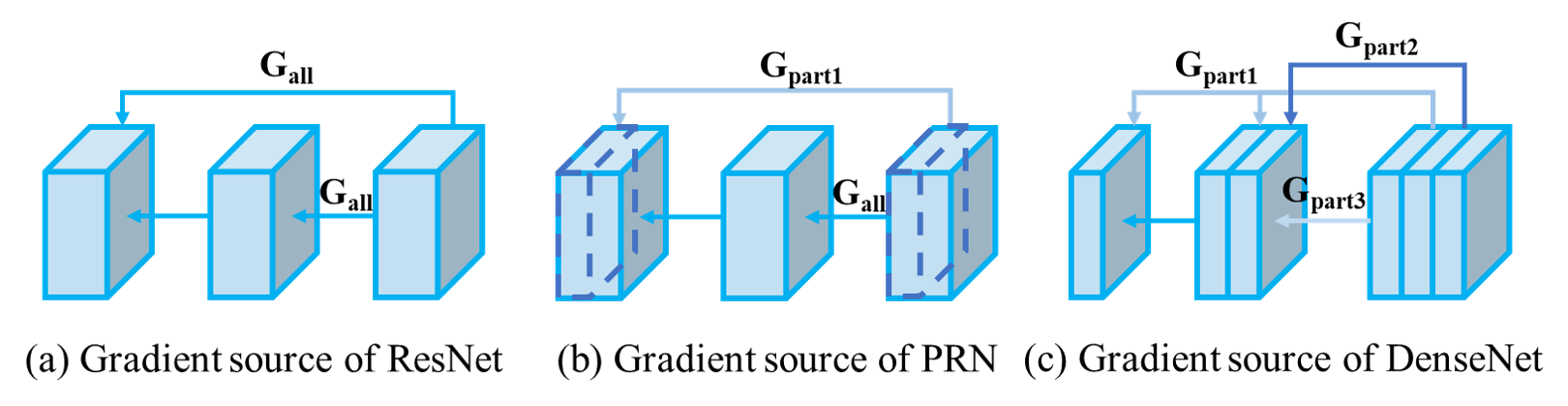}
	\end{center}
	\caption{Gradient source. (a) ResNet, (b) PRN, and (c) DenseNet.}
	\label{fig:source}
\end{figure}

\textbf{Gradient Source:} Figure \ref{fig:source} shows the gradient sources from ResNet \cite{he2016deep}, PRN and DenseNet \cite{huang2017densely} at the first gradient timestamp.  It can be seen from Figure \ref{fig:source} that the concatenation connection-based architectures, such as DenseNet and SparseNet \cite{zhu2018sparsely}, belong to the network that must be specially handled.  This is because in the gradient propagation process, if it is the gradient information propagated by the same layer at a certain gradient timestamp, because the gradient flow has been split beforehand, it cannot be processed like a general network.  As for the residual connection-based architectures, such as ResNet \cite{he2016deep} and PRN, the exact same gradient information is propagated to all layers of outdegree.  Since the outdegree of PRN is only connected to some channels of other layers, it can have a richer combination of gradients than ResNet as a whole.  In addition, there are network architectures that use other split-transform-merge strategies, such as group convolution-based ResNeXt \cite{xie2017aggregated} and depth-wise convolution-based MobileNet \cite{howard2017mobilenets}, etc., which can also increase the number of gradient sources.

\textbf{Summary.} In summary, through the analysis of gradient timestamp and gradient sources generated in the process of gradient backward propagation, we can clearly explain the existing popular network architectures and the information learned by our proposed PRN and the utilization efficiency of parameters.  In ResNet, different layers share many gradients of the same timestamp and the same gradient source, and DenseNet passes the gradient information of the same timestamp but different sources to the corresponding layers.  This part clearly explains why the concatenation connection-based DenseNet can avoid the problem of easily learning a lot of useless information like the residual connection-based ResNet.  Our proposed PRN uses a simple masked residual layer to increase the number of gradient combinations along time axis while maintaining the ResNet network topology, and to divert the gradient sources, thereby increasing the variability of the gradient sources.

\subsection{Analysis of cross stage partial strategy}
\label{sec:acsp}

CSPNet is designed to enhance online learning ability and speed up inference at the same time, so we will discuss the advantages of the CSPNet strategy from these two aspects separately.  In the analysis conducted in Section \ref{sec:agc}, we observed that even if the number of combinations generated by the gradient sources is the same, when the common components received between different combinations are reduced, which makes the gradient components more abundant, and also makes the network learn better.  This phenomenon actually occurs in the process of learning a large number of parameters for singe-layer weights.  For example, dropout \cite{srivastava2014dropout} uses random Bernoulli masking neurons to prevent parameters to learn co-adaptation information.  From a mathematical model point of view, dropout is to update the weights of different parts by using the gradients generated by different inputs, which is equivalent to a random ensemble structure.  As for CSPNet, it directly increases the richness of the gradient combination through the difference in time and the spatial transformation of the gradient on the gradient path.  Next, we will introduce what strategy the CSPNet uses to solve the problem of duplicated gradient information, and how it improves resource usage.

\textbf{Duplicated Gradient Information:} In Section \ref{sec:agc} we analyzed the number of gradient combinations and the effect of diversity on the learning ability of the network.  In CSPNet, we further analyze the gradient information content received by different layers, and design the architecture to improve the efficiency of parameter usage.  From the gradient combination of PRN and SparseNet, it can be found that they have a commonality in the process of increasing the richness of gradient combination, that is, the situation of receiving a large number of duplicated gradient information through residual connection or dense connection is significantly reduced.  We speculate that these duplicated gradients are the main reason for the large number of weights to easily learn the co-adaptation information.  As for PRN, it utilizes gradient timing differences to update the weights of local channels.  With the update process of chain rule, the above timing difference will spread to the entire network, and then achieve a richer gradient combination.  In addition, CSPNet directly uses cross stage connection to make the two paths of the entire stage have a great timing difference, and uses different fusion structures to reduce the duplicated gradient information between stage and stage, or between computational block path and cross stage connection path.

\textbf{Resource Usage Efficiency:} Taking Darknet-53 as an example, suppose that cross stage partial operation divides the feature map into two equal parts according to the direction of the channel.  At this time, the number of input channel and output channel of residual block is halved, while the number of channels in the middle remains unchanged.  According to the above structure, the overall calculation and parameter amount of computational blocks will be reduced to half of the original, and the memory peak is the sum of the size of input feature map and output feature map, so it will be reduced to 2/3 of the original.  In addition, since the input channel and output channel of the convolution layer in the entire computational blocks are equal, the memory access cost at this time will be the smallest.

\begin{table}[h]
	\centering
	\begin{threeparttable}[h]
		\footnotesize
		\caption{Apply CSPNet on different Networks.}
		\label{table:a2}
		\setlength\tabcolsep{4.5pt}
		\begin{tabular}{lccc}
			\toprule
			\textbf{Model} & FLOPs & \#Params & \textbf{Top-1}  \\				
			\midrule
			\textbf{Darknet-53 \cite{redmon2018yolov3}} & 18.57G & 41.57M & 77.2\% \\
			\textbf{+ CSP} & \textbf{13.07G} \textcolor{blue}{(-30\%)} & \textbf{27.61M} \textcolor{blue}{(-34\%)} & \textbf{77.2\%} \textcolor{blue}{(=)} \\		
			\midrule
			\textbf{ResNet-50 \cite{he2016deep}} & 9.74G & 22.73M & 75.8\% \\
			\textbf{+ CSP} & \textbf{8.97G} \textcolor{blue}{(-8\%)} & \textbf{21.57M} \textcolor{blue}{(-5\%)} & \textbf{76.6\%} \textcolor{blue}{(+0.8)} \\		
			\midrule
			\textbf{ResNeXt-50 \cite{xie2017aggregated}} & 10.11G & 22.19M & 77.8\% \\
			\textbf{+ CSP} & \textbf{7.93G} \textcolor{blue}{(-22\%)} & \textbf{20.50M} \textcolor{blue}{(-8\%)} & \textbf{77.9\%} \textcolor{blue}{(+0.1)} \\
			\bottomrule
		\end{tabular}
		\begin{tablenotes}[flushleft]
			\footnotesize
			\item[*] Results are obtained on ImageNet validation set.
		\end{tablenotes}
	\end{threeparttable}
\end{table}

\textbf{Summary.} In summary, CSPNet successfully combines the concept of gradient combination with the efficiency of hardware utilization so that the designed network architecture improves the learning ability and inference speed at the same time.  CSPNet uses only simple channel split, cross stage connection and a small amount of extra transition layers, and successfully completes the preset goal without changing the original network computing units.  Another benefit of the CSPNet is that it can be applied to many popular network architectures and improve overall network efficiency in all aspects.  In Table \ref{table:a2} we show the excellent performance of the CSPNet applied to several popular network architectures.  Finally, because the CSPNet has lower requirements on many hardware resources, it is suitable for high-speed inference on devices with more stringent hardware constraints.

\subsection{Analysis of length of gradient path}
\label{sec:algp}

As discussed in Section \ref{sec:agc}, we understand that the shorter the gradient path of the overall network does not mean the stronger the learning ability.  Furthermore, even if the length of the overall gradient combination path is fixed, we find that the learning ability of the ResNet still degrades when the stacking is very deep.  However, we found that the above problem can be used to disassemble the ResNet into shallower random sub-networks for training during the training phase using stochastic depth \cite{huang2016deep}, which can make the ultra-deep ResNet converge to better results.  The above phenomenon tells us that when analyzing the gradient path, we can not only look at the shortest gradient path and the longest gradient path of the overall network, but need a more detailed gradient path analysis.  In what follows, we will control the gradient path length by adjusting gradient flow during training, and then discuss the gradient length strategy when designing the network architecture from the results.

\begin{figure}[h]
	\begin{center}
		\includegraphics[width=1.\linewidth]{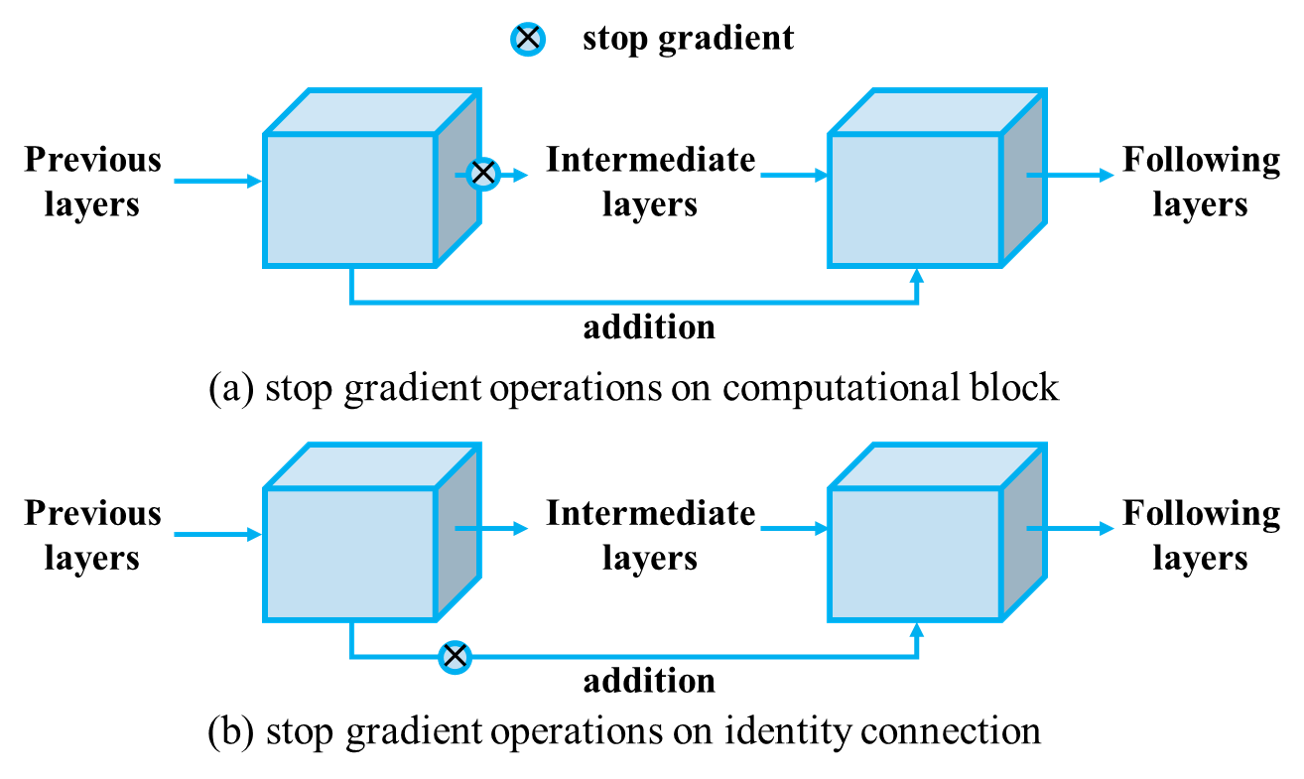}
	\end{center}
	\caption{Architectures for stop gradient ablation studies.}
	\label{fig:stop}
\end{figure}

\begin{table}[h]
\centering
\begin{threeparttable}[h]
	\footnotesize
	\caption{Results of stop gradient ablation study.}
	\label{table:a3}
	\setlength\tabcolsep{4.5pt}
	\begin{tabular}{lccc}
		\toprule
		\textbf{Model} & FLOPs & \textbf{AP$^{box}$} & \textbf{AP$^{mask}$}  \\				
		\midrule
		\textbf{YOLOR-CSP \cite{wang2021you}} & 159.0G & \textbf{51.0\%} & \textbf{41.1\%} \\
		\textbf{YOLOR-CSP \cite{wang2021you} + Figure \ref{fig:stop} (a)} & 159.0G & 48.8\% & 39.5\% \\
		\textbf{YOLOR-CSP \cite{wang2021you} + Figure \ref{fig:stop} (b)} & 159.0G & {47.7\%} & {38.7\%} \\
		\bottomrule
	\end{tabular}
	\begin{tablenotes}[flushleft]
		\footnotesize
		\item[*] Results are obtained on MS COCO validation set.
	\end{tablenotes}
\end{threeparttable}
\end{table}

\textbf{Stop gradient:} First we explore the importance of the shortest gradient length based on ResNet.  Compared to PlainNet, each residual block in ResNet has a part of gradient across the computational block through identity connection in addition to the gradient passing through the computational block.  Here, we perform stop gradient operations on computational block and identity connection respectively, as shown in Figure \ref{fig:stop}.  When we execute stop gradient on identity connection, the gradient path of the overall network will be like PlainNet.  That is to say, the longest gradient path is the same length as the shortest gradient path, and the network depth is also the same.  When we perform stop gradient on a computational block, the shortest gradient path will go directly through all residual connection and directly to the starting layer, and the shortest gradient path length is 1 at this time.  Since each computational block has two layers, its longest gradient path is 2.  We can use these two sets of settings to observe the benefits of residual learning itself and the reduction of gradient path.  We use object detection and instance segmentation in Microsoft COCO dataset as the baseline model to perform ablation study on YOLOR-CSP \cite{wang2021you} and show the results in Table \ref{table:a3}.  Experimental results show that performing a shortened gradient path in ResNet is indeed an important factor for better convergence of deep networks.

\begin{table}[h]
	\centering
	\begin{threeparttable}[h]
		\footnotesize
		\caption{Apply ELAN concept on VoVNet.}
		\label{table:a4}
		\setlength\tabcolsep{4.5pt}
		\begin{tabular}{lccc}
			\toprule
			\textbf{Model} & FLOPs & \textbf{AP$^{box}$} & \textbf{AP$^{mask}$}  \\				
			\midrule
			\textbf{Deep VoVNet \cite{lee2019energy} + ELAN} & 253.4G & 53.3\% & 42.9\% \\
			\textbf{Deep VoVNet \cite{lee2019energy} + ELAN + CSP} & 236.5G & \textbf{53.4\%} & \textbf{42.9\%} \\
			\bottomrule
		\end{tabular}
		\begin{tablenotes}[flushleft]
			\footnotesize
			\item[*] Deep VoVNet is a VoVNet with 99 convolutional layers.
			\item[*] Results are obtained on MS COCO validation set.
		\end{tablenotes}
	\end{threeparttable}
\end{table}

\textbf{Gradient path planning:} From the above analysis and our experiment of model scaling using CSP fusion in YOLOR-CSP, we re-plan the transition layer of VoVNet and conduct experiment.  We first remove the transition layer of each OSA module of the deep VoVNet, leaving only the transition layer of the last OSA module in each stage.  We organize both the longest gradient path of the network and the shortest gradient path through each layer in the same way as described above.  At the same time, we also apply the CSPNet structure to the above network to further observe the versatility of CSPNet, and the related experimental results are shown in Table \ref{table:a4}.  We clearly see that deep VoVNet has changed from failing to converge to one that can converge well and achieve very good accuracy.

\newpage

\textbf{Summary.} In short, from the above experiments and analysis, we infer that when planning the gradient path of the overall network, we should not only consider the shortest gradient path, but should ensure that the shortest gradient path of each layer can effectively been trained.  As for the length of the longest gradient path of the overall network, it will be greater than or equal to the longest gradient path of any layer.  Therefore, when practicing network-level gradient path design strategies, we need to consider the longest shortest gradient path length for all of layers in the network, and the longest gradient path for the overall network.

\section{Experiments}
\label{sec:exp}

\subsection{Experimental setup}
\label{sec:setup}

We use the Microsoft COCO dataset as the basis for performing validation on object detection and instance segmentation.  As for baseline architecture we chose residual-based YOLOv3-SPP \cite{redmon2018yolov3}, and for baseline decoder we chose a combination of YOLOR \cite{wang2021you} and YOLO-v5 (r6.2) \cite{glenn2022yolov5}.  As for baseline training strategy and all methods of training hyperparameters, we follow the rules adopted by YOLOR \cite{wang2021you}.  We name the baseline model trained in the above YOLOR-v3 \cite{wang2021you}.  In the following experiments, we will verify one-by-one the effect of our proposed layer-level, stage-level, and network-level architecture based on the gradient path design strategies.  Finally, we compare the proposed method with baseline-related methods such as YOLOR-v3 \cite{wang2021you} and YOLO-v5 (r6.2) \cite{glenn2022yolov5}.

\subsection{Layer-level gradient path design strategies}
\label{sec:llgp}

\begin{table}[h]
	\centering
	\begin{threeparttable}[h]
		\footnotesize
		\caption{Ablation study of PRN.}
		\label{table:e1}
		\setlength\tabcolsep{4.5pt}
		\begin{tabular}{lccccc}
			\toprule
			\textbf{Model} & FLOPs & \textbf{AP$^{box}$} & \textbf{AP$^{box}_{75}$} & \textbf{AP$^{mask}$} & \textbf{AP$^{mask}_{75}$} \\				
			\midrule
			\textbf{YOLOR-v3 \cite{wang2021you}} & 194.6G & 49.5\% & 53.9\% & 40.9\% & 43.1\%  \\
			\textbf{+ PRN} & 194.6G & \textbf{50.0\%} & \textbf{54.4\%} & \textbf{41.0\%} & \textbf{43.4\%} \\
			\bottomrule
		\end{tabular}
		\begin{tablenotes}[flushleft]
			\footnotesize
			\item[*] YOLO-v3 + PRN equals to YOLOv3-FPRN \cite{wang2019enriching}.
			\item[*] Results are obtained on MS COCO validation set.
		\end{tablenotes}
	\end{threeparttable}
\end{table}

In the experiment of PRN, we set the number of channels shaded by the masked residual layer to half of the original number of channels, and the results obtained in the experiment are shown in Table \ref{table:e1}.  Since the design of PRN maintains all parameters and topology of the entire network, only the addition operation in residual connection is reduced by half, so the overall calculation amount is almost unchanged.  However, YOLOR-PRN gets a significant improvement in accuracy due to the addition of the combination of gradients that each layer uses to update the weights.  Compared to YOLOR-v3, PRN improves 0.5\% AP on object detection, and we can also observe high quality and significant improvement.  On instance segmentation, we improved AP by 0.1\% and AP$_{75}$ by 0.3\%.

\subsection{Stage-level gradient path design strategies}
\label{sec:slgp}

\begin{table}[h]
\centering
\begin{threeparttable}[h]
	\footnotesize
	\caption{Ablation study of CSPNet.}
	\label{table:e2}
	\setlength\tabcolsep{3.0pt}
	\begin{tabular}{lccccc}
		\toprule
		\textbf{Model} & FLOPs & \textbf{AP$^{box}$} & \textbf{AP$^{box}_{75}$} & \textbf{AP$^{mask}$} & \textbf{AP$^{mask}_{75}$} \\				
		\midrule
		\textbf{YOLOR-v3 \cite{wang2021you}} & 194.6G & 49.5\% & 53.9\% & 40.9\% & 43.1\%  \\
		\textbf{+ CSP} & 159.0G & \textbf{51.0\%} & \textbf{55.5\%} & \textbf{41.1\%} & \textbf{43.4\%} \\
		\textbf{+ CSP fusion first} & 158.1G & \textbf{50.8\%} & \textbf{55.3\%} & \textbf{41.0\%} & \textbf{43.3\%} \\
		\textbf{+ CSP fusion last} & 155.6G & \textbf{50.6\%} & \textbf{55.3\%} & \textbf{40.9\%} & \textbf{43.3\%} \\
		\textbf{+ CSP no fusion} & 154.8G & \textbf{50.5\%} & \textbf{55.2\%} & \textbf{40.9\%} & \textbf{43.2\%} \\
		\bottomrule
	\end{tabular}
	\begin{tablenotes}[flushleft]
		\footnotesize
		\item[*] YOLOR-v3 + CSP equals to YOLOR-v4-CSP \cite{wang2021you}.
		\item[*] Results are obtained on MS COCO validation set.
	\end{tablenotes}
\end{threeparttable}
\end{table}

In the CSPNet experiment, we follow the principle of optimizing the inference speed and set the gradient split ratio to 50\%-to-50\%, and we show the experimental results in Table \ref{table:e2}.  Since only half of the channel's feature maps will enter the computational block, we can clearly see that YOLOR-CSP significantly reduces the amount of calculations by 22\% compared to YOLOR-v3.  However, with rich gradient combinations, YOLOR-CSP significantly improves the AP by 1.5\% on the object detection.  Compared to YOLOR-v3, the combination of YOLOR and CSPNet (YOLOR-CSP) added more high-quality results.  We further compare gradient flow truncate operations for reducing repeated gradient information, and we clearly see that the strategy of YOLOR-CSP does learn better than CSP fusion first and CSP fusion last.  It is worth mentioning that no matter what fusion strategy is adopted, the CSP-based architecture has a much lower computational load than YOLOR-v3 and an accuracy far better than YOLOR-v3.

\subsection{Network-level gradient path design strategies}
\label{sec:nlgp}

\begin{table}[h]
\centering
\begin{threeparttable}[h]
	\footnotesize
	\caption{Ablation study of ELAN.}
	\label{table:e3}
	\setlength\tabcolsep{3.0pt}
	\begin{tabular}{lccccc}
		\toprule
		\textbf{Model} & FLOPs & \textbf{AP$^{box}$} & \textbf{AP$^{box}_{75}$} & \textbf{AP$^{mask}$} & \textbf{AP$^{mask}_{75}$} \\				
		\midrule
		\textbf{YOLOR-v3 \cite{wang2021you}} & 194.6G & 49.5\% & 53.9\% & 40.9\% & 43.1\%  \\
		\textbf{+ ELAN-\{1,1\}s} & 126.4G & \textbf{50.2\%} & \textbf{54.5\%} & {40.6\%} & {42.9\%} \\
		\textbf{+ ELAN-\{2,1\}s} & 143.2G & \textbf{51.4\%} & \textbf{55.8\%} & \textbf{41.5\%} & \textbf{43.7\%} \\
		\textbf{+ ELAN-\{2,2\}s} & 164.0G & \textbf{51.8\%} & \textbf{56.5\%} & \textbf{41.6\%} & \textbf{43.3\%} \\
		\bottomrule
	\end{tabular}
	\begin{tablenotes}[flushleft]
		\footnotesize
		\item[*] \{$a$,$b$\}s means stack in computational blocks $a$ and $b$ times on backbone and neck, respectively.
		\item[*] Results are obtained on MS COCO validation set.
	\end{tablenotes}
\end{threeparttable}
\end{table}

In the ELAN experiment, we test the stacking times of computational blocks in backbone and neck respectively, and we show the results in Table \ref{table:e3}.  From this table, we can clearly see that ELAN can still improve the performance of object detection by 0.7\% AP under 35\% less amount of computation than YOLOR-v3.  In ELAN, we can flexibly set the number of stacks to make a trade-off between accuracy and computation.  From the experimental results listed in Table \ref{table:e3}, we can see that under the stack setting of {2,1}s, YOLOR-ELAN can significantly improve the performance of object detection and instance segmentation by 1.9\% AP and 0.6\% AP, respectively, under the condition of reducing the amount of computation by 26\%. 

\subsection{Comparison}
\label{sec:comp}

\begin{table}[h]
\centering
\begin{threeparttable}[h]
	\footnotesize
	\caption{Compariso with baseline methods.}
	\label{table:e4}
	\setlength\tabcolsep{3.0pt}
	\begin{tabular}{lcccc}
		\toprule
		\textbf{Model} & FLOPs & \#Params & \textbf{AP$^{box}$} & \textbf{AP$^{mask}$} \\				
		\midrule
		\textbf{YOLO-v5l (r6.2) \cite{glenn2022yolov5}} & 147.7G & 47.9M & 49.1\% & 40.0\%  \\
		\textbf{YOLO-v5x (r6.2) \cite{glenn2022yolov5}} & 265.7G & 88.8M & 50.9\% & 41.4\%  \\
		\textbf{YOLOR-v3 \cite{wang2021you}} & 194.6G & 64.3M & 49.5\% & 40.9\%  \\
		\textbf{YOLOR-PRN} & 194.6G & 64.3M & {50.0\%} & {41.0\%} \\
		\textbf{YOLOR-CSP} & 159.0G & 54.3M & {51.0\%} & {41.1\%} \\
		\textbf{YOLOR-ELAN} & \textbf{143.2G} & \textbf{34.5M} & \textbf{51.4\%} & \textbf{41.5\%} \\
		\bottomrule
	\end{tabular}
	\begin{tablenotes}[flushleft]
		\footnotesize
		\item[*] Results are obtained on MS COCO validation set.
	\end{tablenotes}
\end{threeparttable}
\end{table}

Finally, we comprehensively compare the three proposed methods, that is, YOLOR-PRN designed by layer-level design strategies, YOLOR-CSP designed by stage-level design strategies, and YOLOR-ELAN designed by network-level design strategies, with baseline YOLOR-v3 and YOLOv5 (r6.2), and the results are shown in Table \ref{table:e4}.  From the table, we see that the model designed based on gradient path design strategy outperforms the baseline-based methods in all aspects.  In addition, regardless of the amount of computation. The amount of parameters, and the accuracy, the YOLOR-ELAN designed by network-level design strategy can obtain the most outstanding performance in an all-round way.  From the results we confirm that based on the gradient path analysis, we are able to devise better network architecture design strategies.  If compared with general data path-based strategies, the architecture designed by data path strategy usually requires additional parameter or computational cost to achieve better accuracy.  In contrast, the three proposed architectures based on gradient path design strategy can significantly improve the overall performance.

\section{Conclusions}
\label{sec:concl}

In this paper we propose a strategy for designing network architectures with gradient paths.  We propose three different gradient path design strategies and these strategies confirm that no matter designing from layer-level, stage-level, or network-level, it can effectively and comprehensively improve the network architecture to achieve great learning ability.  Compared with data path-based design strategies, data path-based strategy often needs to design additional computing units and complex topology to achieve better learning results.  As for gradient path design strategies, it can completely rely on the existing computing units, and re-planning through a simple gradient path can reduce the amount of parameters, computing, hardware resources, and improve the inference speed and network learning effect simultaneously.  In this paper we redefine the strategy for designing a network and create an effective and concise architectural design rule.

\newpage

\section{Acknowledgements}
\label{sec:ack}

The authors wish to thank National Center for High-performance Computing (NCHC) for providing computational and storage resources.


\end{document}